\documentclass[sigconf]{acmart}
\usepackage{multirow}
\usepackage{xcolor}
\usepackage{enumitem}
\usepackage{soul}
\usepackage{url}
\usepackage{subcaption}
\usepackage{pifont}
\usepackage{balance}

\copyrightyear{2024}
\acmYear{2024}
\setcopyright{rightsretained}
\acmConference[KDD '24]{Proceedings of the 30th ACM SIGKDD Conference on Knowledge Discovery and Data Mining}{August 25--29, 2024}{Barcelona, Spain}
\acmBooktitle{Proceedings of the 30th ACM SIGKDD Conference on Knowledge Discovery and Data Mining (KDD '24), August 25--29, 2024, Barcelona, Spain}\acmDOI{10.1145/3637528.3671938}
\acmISBN{979-8-4007-0490-1/24/08}

\begin{document}

\title{CoLiDR: \underline{Co}ncept  \underline{L}earn\underline{i}ng using  Aggregated   \underline{D}isentangled   \underline{R}epresentations}

\author{Sanchit Sinha}
\email{sanchit@virginia.edu}
\affiliation{%
  \institution{University of Virginia}
  \city{Charlottesville}
  \state{VA}
  \country{USA}
}
\author{Guangzhi Xiong}
\email{hhu4zu@virginia.edu}
\affiliation{%
  \institution{University of Virginia}
  \city{Charlottesville}
  \state{VA}
  \country{USA}
}
\author{Aidong Zhang}
\email{aidong@virginia.edu}
\affiliation{%
  \institution{University of Virginia}
  \city{Charlottesville}
  \state{VA}
  \country{USA}
}

\begin{abstract}
Interpretability of Deep Neural Networks using concept-based models offers a promising way to explain model behavior through human-understandable concepts. A parallel line of research focuses on disentangling the data distribution into its underlying generative factors, in turn explaining the data generation process. While both directions have received extensive attention, little work has been done on explaining concepts in terms of generative factors to unify mathematically disentangled representations and human-understandable concepts as an explanation for downstream tasks.
In this paper, we propose a novel method CoLiDR - which utilizes a disentangled representation learning setup for learning mutually independent generative factors and subsequently learns to aggregate the said representations into human-understandable concepts using a novel aggregation/decomposition module. Experiments are conducted on datasets with both known and unknown latent generative factors. Our method successfully aggregates disentangled generative factors into concepts while maintaining parity with state-of-the-art concept-based approaches. Quantitative and visual analysis of the learned aggregation procedure demonstrates the advantages of our work compared to commonly used concept-based models over four challenging datasets. Lastly, our work is generalizable to an arbitrary number of concepts and generative factors - making it flexible enough to be suitable for various types of data.  

\end{abstract}


\maketitle


\section{Introduction}
\label{sec:intro}
The increasing proliferation of Deep Neural Networks (DNNs) has revolutionized multiple diverse fields of research such as vision, speech, and language \cite{he2016deep,vaswani2017attention}. Given the black-box nature of DNNs, explaining DNN predictions has been an active field of research that attempts to impart transparency and trustworthiness in their decision-making processes. Recent research has categorized explainability into progressively increasing levels of granularity. The most fine-grained approaches attempt to assign importance scores to the raw features (e.g. pixels) extracted from the data, while less granular approaches assign importance scores to data points (sets of features). Explaining DNNs using concepts provides the highest level of abstraction, as concepts are high-level entities shared among multiple similar data points that are aligned with human understanding of the task at hand. This makes concept explanation much more global in nature. Many recent approaches for concept-based explanations have attempted to either 1) infer concepts post-hoc from trained models \cite{kim2018interpretability} or 2) design inherently explainable \textit{concept-based models} \cite{alvarez2018towards}, such as the concept bottleneck model (CBM) \cite{koh2020concept}.

A parallel field of research on disentanglement representation learning \cite{bengio2013representation,higgins2018beta,kim2018disentangling,chen2018isolating} attempts to learn a low-dimensional data representation where each dimension independently represents a distinct property of the data distribution. These approaches learn mutually independent generative factors of data by estimating their probability distribution from observed data. Once the probability distribution of the generative factors is estimated, a given sample can be theoretically decomposed and re-generated from its generative factors. Due to their ability to uncover the underlying generative factors, disentanglement approaches are considered highly interpretable.

\begin{figure}[h]
    \centering
    \includegraphics[width=0.35\textwidth]{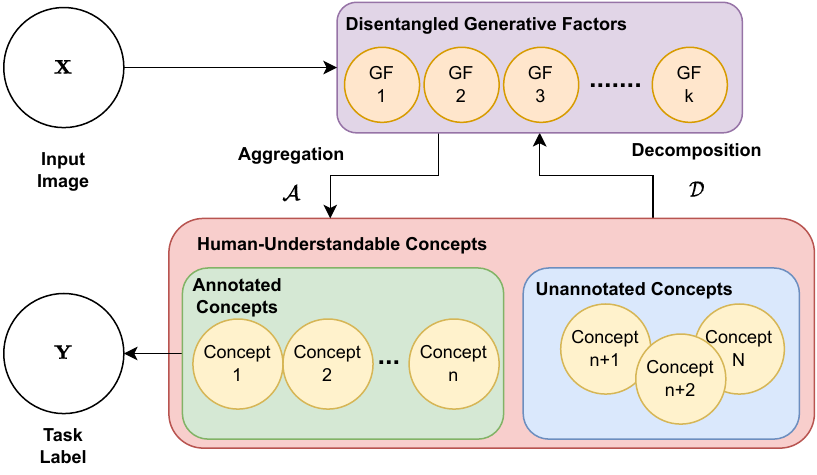}
    \caption{Schematic overview of the proposed CoLiDR approach. The input data distribution $\mathbf{X}$ is first disentangled into $k$ mutually independent generative factors (GFs). Subsequently, the GFs are aggregated into concepts. Note that the concepts are modeled as a set of annotated concepts corresponding to concepts with annotation by humans and a separate set of concepts that are useful for prediction, but are unannotated. Finally, the concepts are utilized for predicting the task label $\mathbf{Y}$.}
    \label{fig:causal}
\end{figure}

Present state-of-the-art approaches do not effectively unify disentangled representation learning with concept-based approaches. Approaches like GlanceNets \cite{marconato2022glancenets} attempt to align concepts with disentanglement with strong assumptions, which are not valid for real-world datasets.
To address the above issues, in this paper, we propose Concept Learning using Aggregated Disentangled Representations (CoLiDR), a self-interpretable approach that combines disentangled representation learning with concept-based explainability. Specifically, CoLiDR learns disentangled generative factors using a disentangled representation learning module, followed by the aggregation of learned disentangled representations into human-understandable concepts using a novel aggregation/decomposition module and subsequently a task prediction module that maps concepts to task labels. Our experiments show that interventions on learned concept representations can fix wrongly classified samples, which makes CoLiDR useful for model debugging. Figure~\ref{fig:causal} demonstrates the schematic overview of CoLiDR, where the learned disentangled generative factors from data are aggregated into concepts, which are further utilized for task prediction. Specifically, our main contributions are as follows: 1) Propose CoLiDR framework, a novel method to aggregate disentangled representations into human understandable concepts, while achieving at-par performance to other commonly utilized concept-based models such as CBMs and GlanceNets, 2) Improve model debugability by test-time interventions, and 3) Provide a flexible framework which can be generalized to arbitrary number of generative factors and concepts.

\section{Related Work}
\label{sec:related}

\subsection{Related Work on Disentangled Representation Learning}

Disentangled representation learning has long been a fascinating study aimed at separating distinct informational factors of variations in real-world data \citep{bengio2013representation,locatello2019challenging}. 
Due to its probabilistic framework and flexibility to customize training objectives, the variational autoencoder (VAE) \citep{kingma2014auto} is a commonly used architecture in the study of disentangled representation learning, which can capture different factors of variation with its encoder and decoder. Based on the traditional VAE, \citet{higgins2018beta} proposed a variant, $\beta$-VAE, which introduces an additional hyperparameter to scale the importance of the regularization term. By changing the weight of the regularization term, $\beta$-VAE can control the trade-off between the reconstruction of inputs and the disentanglement of latent variables. Instead of optimizing the Kullback–Leibler (KL) divergence between the latent distribution and standard Gaussian prior for disentanglement as done by VAE and $\beta$-VAE, FactorVAE \citep{kim2018disentangling} and $\beta$-TCVAE \citep{chen2018isolating} further decompose the regularization term and propose to directly penalize the total correlation between latent variables, which are shown to better disentangle the variables.

While early studies in disentanglement representation learning attempted to learn independent latent variables by modifying the training objective of VAE \cite{higgins2018beta,kim2018disentangling,chen2018isolating}, \citet{locatello2019challenging} showed that it is impossible to learn identifiable disentangled latent variables without any supervision. A series of subsequent studies were then proposed to learn disentangled representations with better identifiability using different kinds of supervision \citep{locatello2020disentangling,shu2020weakly,shen2022weakly}.   


\subsection{Related Work on Concept Based Explanations}

Due to the black-box nature of deep learning models, various approaches have been explored to provide explanations for the outcome of deep neural networks \citep{koh2020concept,pedapati2020learning,jeyakumar2020can,heskes2020causal,o2020generative}. One important direction is to interpret the models with intuitive and human-understandable concepts, which are usually high-level abstractions of the input features. Various attempts have been made to automatically learn the concepts for different tasks \citep{kim2018interpretability,ghorbani2019towards,yeh2020completeness,wu2020towards,goyal2019explaining}. Among them, the concept bottleneck model (CBM) \citep{koh2020concept} is a commonly used approach to incorporate the learning of concepts into deep neural networks. By constructing a low-dimensional intermediate layer in the models, CBMs are able to capture the high-level concepts that are related to the downstream tasks. Numerous studies have been carried out to adapt CBMs for tasks in various domains \citep{sawada2022concept,jeyakumar2021automatic,pittino2021hierarchical,bahadori2020debiasing,marconato2022glancenets}. The most relevant works to our proposed approach are Concept Bottleneck Models \cite{koh2020concept}, Concept Embedding Models \cite{zarlenga2022concept}, CLAP \cite{taeb2022provable} and GlanceNets \cite{marconato2022glancenets}, details of which are discussed in the next Section.

\section{Unifying Concept-based Modeling with Variational Inference}
\label{sec:background}
\noindent \textbf{Problem setup.} The concept learning problem is characterized as a two-step process \cite{koh2020concept} for a given training sample $\{\mathbf{x},\mathbf{y},\mathbf{C}\}$. Given an observation $x$, a concept-based model learns a function to map the input to its associated human-understandable concepts $\mathbf{C} = \{c_1,\cdots,c_N\}$. Subsequently, a predictor will project the concept embeddings to $\hat{y}$, which is the prediction for the label $y$ of the sample in a downstream task. 

\noindent \textbf{Assumptions.} Following previous work in concept-based models \cite{koh2020concept,marconato2022glancenets}, concept models follow two fundamental assumptions as follows: 1) the data distribution in the input space can be accurately mapped to the distributions of concepts in the latent space, and 2) the concept scores are necessary and sufficient to predict labels of samples in downstream tasks.

\subsection{Variational Inference-based Concept-learning for Greater Interpretablity}
Utilizing disentanglement approaches to explain the data generation process has been well studied recently \cite{kingma2014auto,higgins2018beta}. A typical disentanglement system attempts to learn a fundamental representation of mutually independent \textit{generative factors (GFs)} usually modeled as random variables from independent probability distributions. The Variational Inference (VI) process utilizes the learned generative factors to emulate the data generation procedure. Even though variational inference has been successfully utilized for controlled data generation with success \cite{creswell2018generative,higgins2018beta}, utilizing them to improve the interpretability of concept-based models is still relatively underexplored. Our work attempts to fill this gap. Below, we list the advantages of using variational inference-based approaches for concept learning: 
\begin{itemize} [leftmargin=*, parsep=0pt, itemsep=0pt, topsep=0pt]
    \item \textbf{Provides a controlled data generation process:} In addition to emulating the data generation process, VIs offer control via interventions over GFs - a desirable property for stakeholders of a concept-based model. 
    \item \textbf{Captures the continuous space mappings:} Several works \cite{margeloiu2021concept,sinha2023understanding} have demonstrated that learning discrete mapping of input samples to concepts is susceptible to fragility. VI smoothens both the input and concept spaces to learn a continuous, well-trained and robust mapping function. 
    \item \textbf{Easier to isolate confounds:} Another well-documented problem in concept-based models is the encoding of non-related confounds in the concept representations \cite{yuksekgonul2022post}. VI methods are successful in isolating confounds - hence reducing the effect of encoding noise, spurious correlations, etc. in the concept representations.
\end{itemize}

\begin{figure}
    \centering
    \includegraphics[width=0.4\textwidth]{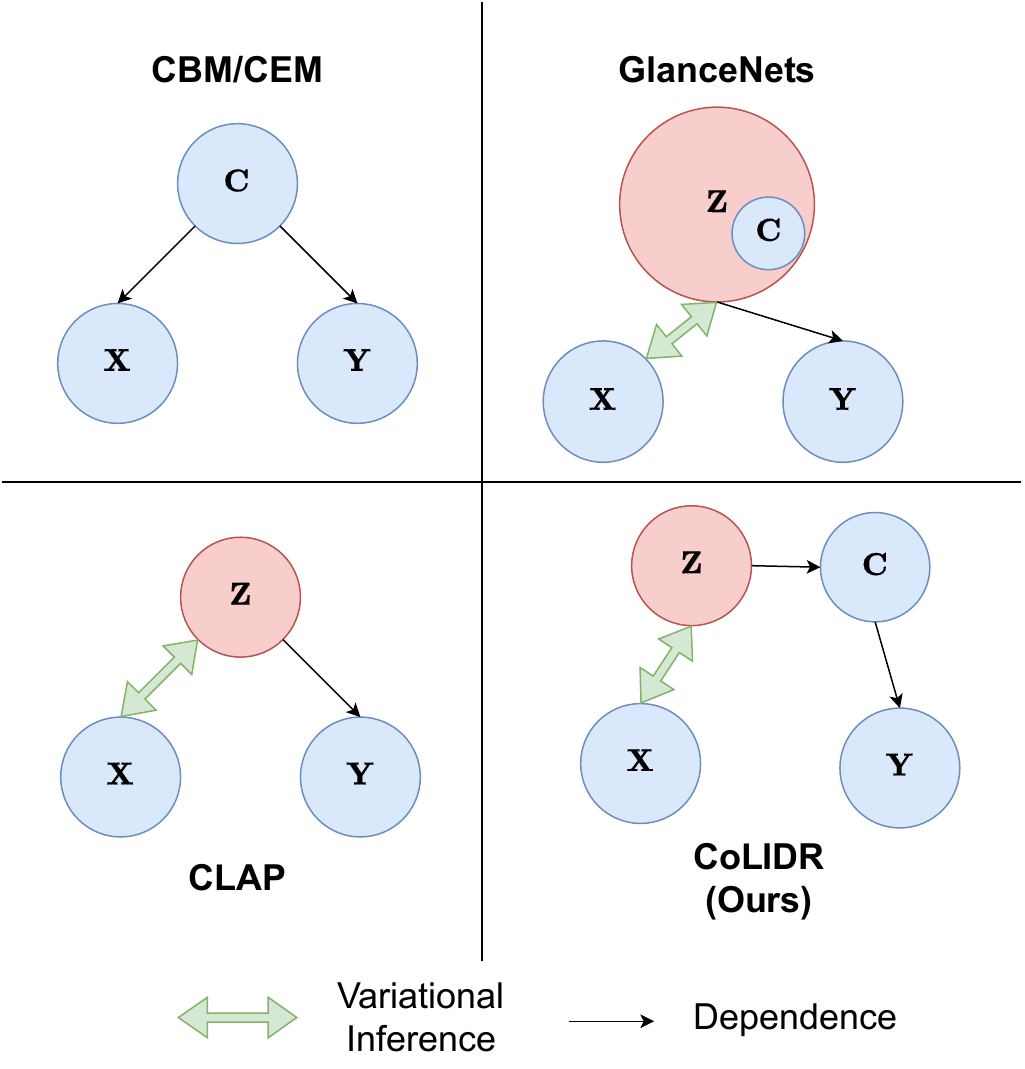}
    \caption{A schematic view of the underlying assumptions considered across SOTA concept-based models, CBM/CEM, CLAP, GlanceNet, and CoLiDR. The circles in blue represent directly observable attributes, input sample $X$, task label $Y$ and representative concepts $C$. The red circles represent learned representations.}
    \label{fig:compare-process}
\end{figure}


\subsection{Comparison with Existing Approaches}
\label{subsec:compare}
We begin the discussion by presenting an \textit{anti-causal} \cite{kilbertus2018generalization} model to visualize the data generation process of our proposed approach and compare it to multiple comparable recent state-of-the-art approaches in Figure~\ref{fig:compare-process}. Note that we represent the task labels as $\mathbf{Y}$, the set of concepts as $\mathbf{C}$, the disentangled generative factors as $\mathbf{Z}$, and the data distribution as $\mathbf{X}$. The single-edge arrows model the dependency between distributions while the double-ended arrows represent variational inference - modeling the generative process.

\noindent {\bf Comparisons to CBM/CEM.}
Concept Bottleneck Models \cite{koh2020concept} and Concept Embedding Models \cite{zarlenga2022concept} do not incorporate variational inference. Nevertheless, they visualize the data-generative process as being conditioned on the annotated concepts and also assume that the task labels are entirely conditioned on the concepts.

\noindent {\bf Comparisons to CLAP.} CLAP \cite{taeb2022provable} is one of the first works utilizing variational inference to model the data generation process. CLAP considers a part of the disentangled space representing relevant GFs to be conditioned on the task labels and a part of it representing confounds to be conditioned on independent normal distributions. However, CLAP does not utilize human-annotated concepts - making the learned disentangled space $\mathbf{Z}$ unidentifiable (Refer \cite{locatello2020disentangling}). This makes CLAP not comparable to our approach.

\noindent {\bf Comparisons to GlanceNets.}
GlanceNets \cite{marconato2022glancenets}, to our knowledge, is the only existing approach that attempts to bridge the gap between VI and concept learning. The GlanceNets approach incorporates a Variational Autoencoder based disentangling mechanism to learn the generative factors (GFs) of the distribution and formulate the concept learning problem as a one-to-one mapping between a subset of generative factors (or disentangled representations) and human-understandable concepts. As shown in Figure~\ref{fig:compare-process}, the annotated concepts directly supervise a part of the learned disentangled space $\mathbf{Z}$. However, in doing so it makes an implicit assumption that the \textbf{human-understandable concepts act as generative factors} of the data distribution. This is a strong assumption for two distinct reasons. Firstly, multiple generative factors can contribute as a constituent of a particular concept and multiple concepts can share the same generative factors. Secondly, human-annotated concepts are by definition, \textit{abstract and high-level} and do not necessarily model the fine-grained data generation process. Utilizing abstract concepts to supervise the disentangling process may undermine the disentanglement procedure, as disentangled representations are supposed to capture the \textbf{low-level} GFs of the data distribution instead of high-level abstract human concepts. The model is shown to be effective on two synthetic datasets, dSprites and MPI-3D, which are procedurally generated using carefully curated GFs that correspond to human-understandable concepts. In addition, CoLiDR can effectively generalize to datasets where GFs are completely unknown.
\noindent \textbf{Unannotated concepts.} Note that we model the concept set as a union of known, annotated concepts, and unannotated concepts. This modeling approach is common in unsupervised concept learning approaches like \cite{alvarez2018towards}, but has not yet been studied for Variational Inference based concept learning. The assumption is that there exists some concepts relevant to the task prediction but are not manually annotated. 

Table~\ref{tab:baseline-summary} lists the differences between the discussed approaches based on - (i) the Presence of VI, (ii) the Presence of unsupervised GFs, (iii) the Presence of supervised concepts, and (iv) the Presence of unannotated concepts (Discussed in Methodology). 

\begin{table}[h]
\centering
\resizebox{0.49\textwidth}{!}{
\begin{tabular}{ccccc}
\hline

Method   &  Variational \newline  & Unsupervised  & Supervised & Unannotated  \\

& Inference & GF & Concepts & Concepts \\

\hline

CBMs/CEMs      &      \ding{55}  &    \ding{55}     &  \ding{55}     &   \ding{55}    \\
GlanceNet      &     \textcolor{red}{\ding{52}}   &  \ding{55}*   &    \textcolor{red}{\ding{52}}   & \ding{55}      \\

CLAP     &     \textcolor{red}{\ding{52}}    &   \textcolor{red}{\ding{52}}    &   \ding{55}     &  \textcolor{red}{\ding{52}}    \\

CoLiDR      &    \textcolor{red}{\ding{52}}    &   \textcolor{red}{\ding{52}}   &  \textcolor{red}{\ding{52}}  &   \textcolor{red}{\ding{52}}    \\

\hline                                                
\end{tabular}
}
\caption{A summary of salient features of our method as compared to the baselines. The asterisk represents partial supervision over the generative factors by GlanceNets.}
\label{tab:baseline-summary}
\end{table}

\subsection{Drawbacks of Supervising GFs with Concepts}
Although it might be tempting to utilize concepts as GFs like what GlanceNets \cite{marconato2022glancenets} did, it is dangerous for a variety of reasons. In this section, we provide the drawbacks associated with this line of thought with a concrete and practical example. Let us consider as a sample from the CelebA dataset \cite{liu2015deep} which consists of facial photographs of celebrities. Each image in the dataset is annotated with binary concepts such as the presence of ``black hair'', ``blonde hair'', ``wavy hair'', ``straight hair'' which are easily understandable by humans - implying they are \textit{abstract}. Consider the assumption (fundamental in \cite{marconato2022glancenets}) that each of these concepts also align with the generative factors of the data distribution as well.  

\noindent {\bf Limited concept abstraction.}
As it can be seen, the assumption is flawed - two identical images with straight hair and different colors (Black and Blonde) would be sampled from completely different distributions despite sharing the same \textit{type} of hair. Images of two people having the same hair color but with wavy and straight hair respectively would also be sampled from completely different distributions despite sharing the same color. Hence, there is no fine-grained control over the actual generative factors. Furthermore, as concepts such as ``attractiveness'' are very subjective and contain a large number of constituent underlying distributions, it is much more effective to understand constituent distributions than directly aligning attractiveness with a single disentangled representation. 

\noindent {\bf Limited intra-concept hierarchy modeling.}
By definition of disentanglement, generative factors and concepts would be independent of each other. Any hierarchical relationship cannot be encoded in such a space. For example, the concept ``black hair'' is ideally sampled from {mutually independent distributions representing ``blackness''} and multiple other distributions constituting in construction of hair themselves. As such, ``blackness'' can also be combined with distributions modeling facial hairs for ``beard'' concepts. 

Hence, instead of directly aligning the concepts with disentangled representations, it is more explainable and reasonable to understand the actual distributions constituting a concept.  Relaxing the assumption of aligning disentangled latent space representations with concepts helps us to better capture and explain constituent distributions in each concept - in turn improving explainability.

\section{Methodology}
\label{sec:method}

\begin{figure*}[h]
    \centering
    \includegraphics[width=\textwidth]{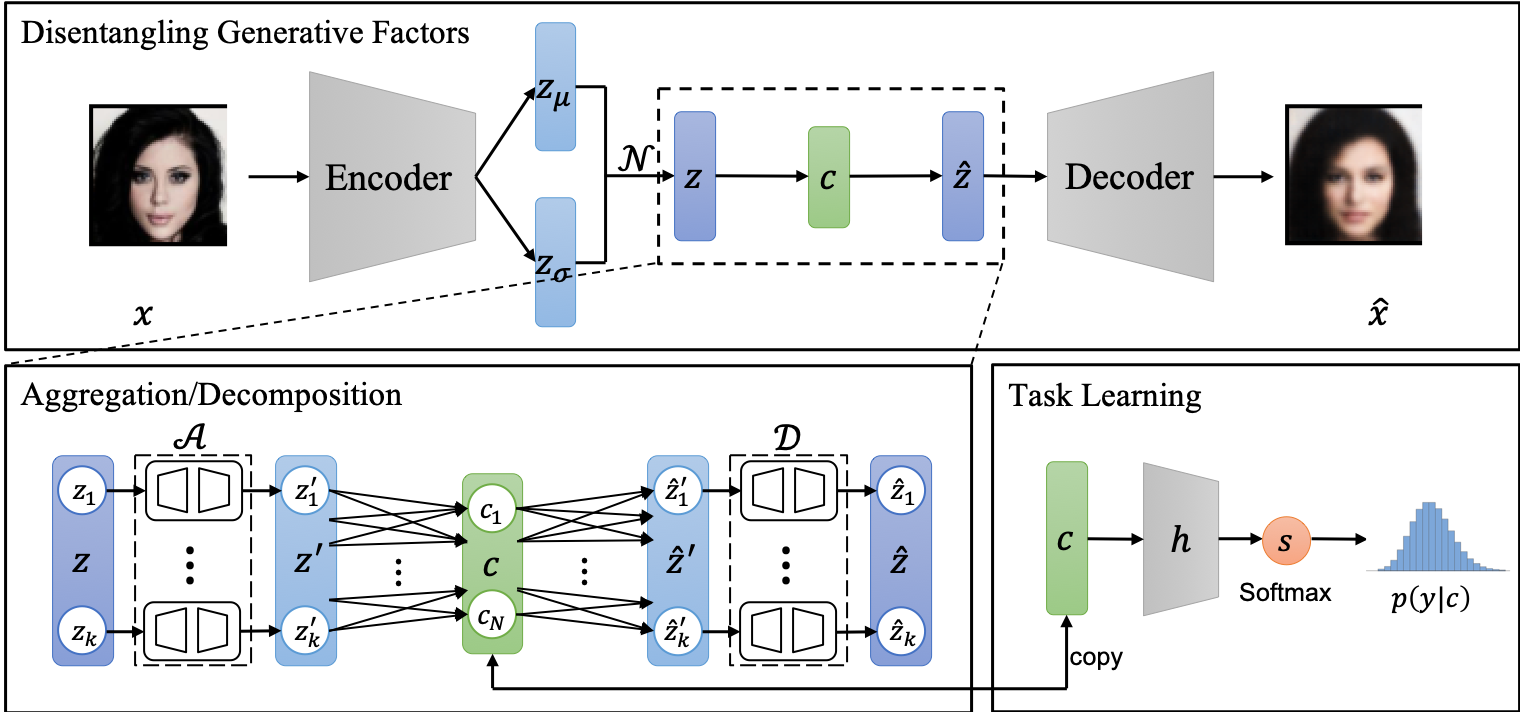}
    \caption{Architecture of the proposed CoLiDR approach. CoLiDR consists of three modules - the Disentangled Representations Learning (DRL) Module which learns disentangled generative factors (top), the Aggregation/Decomposition Module which learns to aggregate the generative factors into concepts and subsequently decompose them back into generative factors (bottom-left) and the Task Learning module (bottom-right) which utilizes the concepts to perform task label prediction.}
    \label{fig:method}
\end{figure*}
We first provide a broad overview of the proposed approach CoLiDR in Section~\ref{subsec:overview}. Section~\ref{subsec:ldr} details the architectures utilized for computing disentangled representations. Subsequently, Section~\ref{subsec:agg-decomp} introduces the Aggregation/Decomposition module and the task prediction network. Finally, Section~\ref{subsec:training} describes the training procedure and additional Disentangled Representation (DR) Consistency loss which regularizes the Aggregation/Decomposition module to correctly learn representations to concept mapping.

\subsection{Overview of CoLiDR}
\label{subsec:overview}
CoLiDR consists of three distinct modules as shown in Figure~\ref{fig:method}. The first module learns the disentangled generative factors from the input data. The second module aggregates the disentangled representations and maps them to human-understandable concepts. The third module maps the concepts to the task label.

\subsection{Disentangled Representations Learning (DRL) Module}
\label{subsec:ldr}
The first step of the proposed approach involves learning disentangled representations corresponding to the various generative factors in the data. As depicted in {Figure~\ref{fig:causal}}, the generative factors form the basis of the data generation process itself. Our DRL module learns the disentangled generative factors that can be used for data generation.
Suppose $z$ is the given embedding of the generative factors and $x$ is the corresponding generated data.
The underlying data generation process $\mathit{p}_\theta (x|z)$ can be estimated using a $\beta$-VAE, which estimates the maximum likelihood using variational inference. 
Following \cite{kingma2014auto}, we maximize the Evidence Lower Bound (ELBO) to model the posterior distribution $q_\phi(z|x)$ and the distribution of data generation $p_\theta(x|z)$ as detailed below. 
\begin{equation}\small
    ELBO = - \beta~\mathbb{D}_{KL}(q_{\phi}(\mathbf{z}|\mathbf{x})~||~p(\mathbf{z})) + \mathbb{E}_{q_{\phi}(z|x)}[log ~p_{\theta}(\mathbf{x}|\mathbf{z})] 
    \label{eq:elbo}
\end{equation}
The $\mathbb{E}$ term in Formula~\ref{eq:elbo} comprises the reconstruction loss between the input $x$ and predicted reconstruction $\hat{x}$, usually taken as Mean Square Error (MSE). The prior distribution $\mathit{p}$ is usually taken as a standard Gaussian distribution which encourages the covariance matrix of the learned distribution to be diagonal, enforcing independence constrain \cite{higgins2018beta,kingma2014auto}. The tunable hyperparameter $\beta$ acts as a  quantitative measure of the extent of disentanglement.

$\beta$-VAE estimates $q_\phi$ using an encoder that maps $x$ from the input observation space in $\mathbb{R}^d$ to the disentangled representation space in $\mathbb{R}^k$, where $d$ and $k$ are the dimensions of the input and latent space respectively. The distribution $p_\theta$ is estimated using a decoder that maps the disentangled representation space back to the observation space ($\mathbb{R}^k \rightarrow \mathbb{R}^d$). Specifically, the encoder encodes the input as an estimated mean $z_\mu \in \mathbb{R}^k$ and a standard deviation vector $z_\sigma\in \mathbb{R}^k$, from which the latent representation ${z}$ is sampled from the Gaussian multivariate distribution $\mathcal{N}(z_\mu, \text{diag}(z_\sigma^2))$.

\subsection{Aggregation/Decomposition Module}
While the DRL module learns disentangled generative factors in the latent space, the factors may be too fine-grained to be aligned with human understanding. However, these disentangled generative factors can be considered as an independent basis of human-understandable concepts as shown in Figure~\ref{fig:causal}.
\label{subsec:agg-decomp}
\subsubsection{Aggregation of Generative Factors into Concepts}
We propose the Aggregation module which \textit{aggregates} the disentangled underlying generative factors of data into human-understandable concepts. Specifically, given a latent representation ${z} = [z_1,\cdots,z_k]^\top$, each concept $c_i \in \{c_i\}_{i=1}^N$ can be considered as a combination of all disentangled factors $z_1,\cdots,z_k$. 
To increase the expressiveness of our model in learning concepts from the generative factors while keeping each concept as a linear aggregation of different factors for interpretability, we propose to encode each generative factor $z_j (j\in \{1,\cdots, k\})$ with an independent neural network $a_j$, which learns the non-linear mapping from $z_j$ to $z'_j$ for concept learning. Subsequently, our model learns the linear combinations of components in $z' = [z'_1,\cdots,z'_k]^\top$ to represent high-level concepts $c_1,\cdots,c_N$ that are relevant to the downstream task. Given the posterior distribution $q_\phi(z|x)$ in $\beta$-VAE, the learned concepts can be formulated as
\begin{equation}
    c = f(\mathcal{A}(z)), 
\end{equation}
where $z \sim q(z|x)$, $f$ is a one-layer fully connected model, and $\mathcal{A}$ is defined as
\begin{equation}\small
    \mathcal{A}(z) = [z'_1,\cdots,z'_k]^\top = [a_1(z_1), a_2(z_2), \cdots, a_k(z_k)]^\top.
\end{equation}

\subsubsection{Concept Level Supervision}


Given the annotation of $n$ manually defined concepts, we can supervise the learning of the first $n$ concepts to be aligned with human knowledge. Formally, given the estimated scores $c_1,\cdots,c_n$ and human annotations $l_1,\cdots,l_n$, the supervision can be performed by minimizing
\begin{equation}
    \mathcal{L}_{con} = \sum_{i = 1}^n \textsf{BCE}(c_i, l_i),
\end{equation}
where $\textsf{BCE}$ is the Binary Cross Entropy Loss. In addition to the supervised learning of human-annotated concepts, our model is designed to automatically capture the information of the remaining concepts $c_{n+1},c_{n+2}\cdots,c_N$ from the input in an unsupervised manner as it is not possible to annotate a truly closed set of concepts. Specifically, the unsupervised learning of the unannotated concepts is performed through the training of concept decomposition for the reconstruction of generative factors, which are described in Sections \ref{sec:decomposition} and \ref{sec:training}.

\subsubsection{Decomposing Concepts into Generative Factors} \label{sec:decomposition}
Corresponding to the aggregation module, we propose the decomposition module $g$ which learns a mapping from the concept embedding $c = [c_1, \cdots, c_N]^\top$ back to transformed disentangled representations ($\mathbf{z'}$) which are further transformed back to the original disentangled representations $\mathbf{z}$ with decoders $\mathcal{D} = \{d_1,\cdots,d_n\}$. Mathematically, 
\begin{equation}
\begin{aligned}
    \hat{z} &= \mathcal{D}( g(c) ) = [d_1(g(c)),\cdots,d_n(g(c))]^\top\\
\end{aligned}
\end{equation}
where $\hat{z}$ is the estimated generative factors from the given concepts $c$.
Here the decomposition module acts as the inverse of the aggregation module and maps concepts back to the disentangled representations. Instead of using one decoder directly for the mapping from $c$ to $z$, we use independent decoders to project each dimension in $z'$ to the corresponding dimension in $z$, which mitigates the problem of concept leakage from concepts to unrelated generative factors \cite{margeloiu2021concept}.

\subsubsection{Task Prediction using Learned Concepts}
Task prediction entails the mapping from learned concepts in $\mathbb{R}^N$ to the prediction of task labels in $\mathbb{R}^m$, where $m$ is the number of categories in classification tasks. The model prediction can be formulated as
\begin{equation}
    p(y|{c}) = \text{Softmax}(h(c))
\end{equation}
where the function $h$ is a shallow neural network. Although $h$ can be chosen up to task requirements, we utilize a linear polynomial to estimate $h$ for model interpretability, with only human-annotated concepts as the input to see how well they can be used to explain the model prediction:
\begin{equation}
   \mathit{h}({c}) = w_0 + \sum^{n}_{i=1} w_i \cdot c_i,
\end{equation}
where $w_0, w_1, \cdots, w_ns$ are the bias and weights in the predictor. Given the true labels $y$, the loss for the task learning is defined as
\begin{equation}
    \mathcal{L}_{pred} = \Phi\left(p(y|c), y\right),
\end{equation}
where $\Phi$ can be (binary) cross-entropy loss or mean square loss depending on the type of the downstream task.

\subsection{Disentangled Representation Consistency and End-to-end Training} \label{sec:training}
\label{subsec:training}
The three modules of CoLiDR (DRL, Aggregation/Decomposition, Task Learning) are trained end-to-end. To maintain the consistency of learned representations of generative factors, we enforce $z$ and $\hat{z}$ to be as similar as possible. We propose to use a DR consistency loss ($L_{drc}$), which can be formulated as:
\begin{equation}
     \mathcal{L}_{drc} = \|z - \hat{z}\|_2^2
\end{equation}
\noindent For a given set of disentangled generative factors $\mathbf{z}$, there exists a family of \textit{surjective} functions $f$ - which map from $\mathbf{z'}$ to concepts $\mathbf{c}$. Consequently, there exists a family of functions $g$ \textit{inverse coupled} with $f$, which maps from concepts $c$ to $z$. As the function $f$ is subjective, computing a direct inverse is intractable and hence requires enforced consistency through $\mathcal{L}_{drc}$. The function $g$ is NOT a standalone family of functions as they are inverse of a surjective function $f$. 

In addition, we also encourage sparsity on the transformed representations ${z'}$ to identify the \textit{most important} generative factors that are aggregated to compose the concepts and reduce the impact of non-relevant factors. The overall training objective can be given by:
\begin{equation}
    \mathcal{L} = ELBO + \lambda_1 \mathcal{L}_{con} + \lambda_2 \mathcal{L}_{pred} + \lambda_3 \mathcal{L}_{drc} + \lambda_4 \|z'\|_1.
    \label{eq:final-obj}
\end{equation}
\section{Experimental Setup}
\subsection{Dataset Descriptions}
\begin{itemize}[leftmargin=*, parsep=0pt, itemsep=0pt, topsep=0pt]    
\item \textbf{D-Sprites} \cite{dsprites17}: D-Sprites consists of procedurally generated samples from six independent generative factors. 
Each object in the dataset is generated based on two categorical factors (shape, color) and four numerical factors (X position, Y position, orientation, scale).
The six factors are independent of each other. The dataset consists of 737,280 images. We randomly split data into train-test sets in a 70/30 split.
\item \textbf{Shapes3D} \cite{3dshapes18}: Shapes3D consists of synthetically generated samples from six independent generative factors consisting of color (hue) of floor, wall, object (float values) and scale, shape, and orientation in space (integer values). The dataset consists of 480,000 images. We randomly split the data into train and test sets in a 70/30 split.
\item \textbf{CelebA} \cite{liu2015deep}: CelebA consists of about 200,000 178 × 218 sized RGB images of center-aligned facial photographs of celebrities. The faces are annotated with 40 binary concepts like hair color, smile, attractiveness, etc. Some of the features in the set are simple and observable like color (black, blonde) and style of hair (wavy, bangs). However, many concepts are abstract and subjective like attractiveness, heavy makeup, etc. We only consider the objective concepts for experiments. We center-crop the images to 148x148 and subsequently resize them to 64x64. 
\item \textbf{AWA2} \cite{lampert2013attribute}: Animals with Attributes-2 consists of 37,322 images of a combined 50 animal classes with 85 binary concepts like number of legs, presence of tail, etc. We remove certain subjective concepts such as ``eats fish''. AWA2 is neither centered nor cropped and consists of significant background noise, making it significantly harder to disentangle. We resize all images to 64x64 and combine the train and test splits. We use 70\% of the data for training and 30\% for testing.

\end{itemize}

\subsection{Dataset Task Descriptions}
\noindent \textbf{Synthetic datasets:} As dSprites and Shapes3D datasets are procedurally generated, they do not contain an inherent downstream task, hence we construct downstream tasks using combinations of GFs Similar to \cite{marconato2022glancenets} and \cite{raman2023concept}. For each task, we consider two GFs at random and a sample has the label ``$1$'' when all factors satisfy a pre-defined criterion and. For categorical factors, we consider the presence of exact values as Truth, while for continuous factors we use a threshold. More details on task construction can be found in the Appendix.

\noindent \textbf{Real-world datasets:} For the CelebA dataset, the downstream task is cluster assignment \cite{marconato2022glancenets}. For AWA2, the downstream task is classification.

\begin{table*}[t]
\centering
\begin{tabular}{cc|cccc|cccc}
\hline
                              &           & \multicolumn{4}{c|}{Task Accuracy ($\uparrow$)} & \multicolumn{4}{c}{Concept Error ($\downarrow$)} \\ \cline{3-10} 
                              &           & dSprites    & Shapes3D    & CelebA    & AWA2    & dSprites     & Shapes3D    & CelebA    & AWA2    \\ \hline
\multicolumn{1}{c|}{wo/ dis.} & CBM/CEM       & 0.924       & 0.913       & 0.828     & 0.531   & 0.068        & 0.0910      & 0.0378    & 0.33    \\ \hline
\multicolumn{1}{c|}{}         & GlanceNet & 0.931       & 0.915       & 0.814     & 0.517   & 0.070        & 0.0910      & 0.0379    & 0.44    \\
\multicolumn{1}{c|}{w/ dis.}  & CoLiDR - VAE    & \bf 0.931       & 0.911       & 0.821     & 0.515   & 0.070        & 0.0912      & 0.0377    & 0.39    \\
\multicolumn{1}{c|}{}         & CoLiDR - $\beta$-VAE (no $\mathcal{L}_{drc}$)    &   0.921     &   0.911     &  0.819    &  0.515  & 0.071  &  0.0919  &   0.0381  & 0.39     \\ 
\multicolumn{1}{c|}{}         & CoLiDR - $\beta$-VAE    & 0.929       & \bf 0.916       & \bf 0.828     & \bf 0.521   & \bf 0.069        & \bf 0.0910      & \bf 0.0375    & \bf 0.36    \\ \hline
\end{tabular}
\caption{Average task accuracy and concept errors across four datasets for concept-based models with disentanglement learning (GlanceNet, CoLiDR - VAE, CoLiDR - $\beta$-VAE) and models without disentanglement learning (CBM). ``CoLiDR - VAE'' is a version of our model with a vanilla VAE (parameterized by $\beta=1$) while ``CoLiDR - $\beta$-VAE'' is another version with a $\beta$-VAE (parameterized by various $\beta$s) discussed in Appendix. Concept Errors are reported for dSprites and Shapes3D as RMSE while for CelebA and AWA2 as 0-1 error. Best results for models with disentanglement learning are marked in \textbf{bold}.}
\label{tab:performance-compare}
\end{table*}

\subsection{Model Implementation Details}
\noindent \textbf{Disentangled representations learning (DRL) module}. 
We utilize two different Variational Autoencoder architectures for estimating disentangle representations in the DRL module. We utilize a standard VAE \cite{kingma2014auto} and a $\beta$-VAE \cite{higgins2018beta}. Even though similar in formulation, a $\beta$-VAE is parameterized by a tunable hyperparameter $\beta$ which controls the \textit{strength} of disentanglement. For both VAEs, the encoder is a 5-layer CNN with BatchNorm and LeakyReLU as the activation function. The decoder is modeled symmetrically to the encoder with five Transpose Convolutional Layers. The size of the latent space $k$ is set as 64 for d-Sprites and Shapes3D and 512 for CelebA and AWA2 datasets.

\textbf{Aggregation/Decomposition module}.
The Aggregation module is composed of the set of transformation operations $\mathcal{A}$ and mapping function between transformed representations and concepts $f$. We model each of the $k$ neural networks $a_i$ for $i\in\{0,1,..,k\}$ as a 3-layer network. For dSprites and Shapes3D the network consists of layers sized [64,64,1] and for CelebA and AWA2 the network consists of layers sized [512,512,1] with ReLU as the activation function. The function $f$ is modeled as a single fully connected layer.
Similarly, the Decomposition module is composed of the set of inverse transformation operations $\mathcal{D}$ and mapping function between concepts and transformed representations $g$. We model each of the $k$ neural networks $d_i$ for $i\in\{0,1,..,k\}$ as a 3-layer network.  For dSprites and Shapes3D the network consists of layers sized [1,64,64] and for CelebA and AWA2 the network consists of layers sized [1,512,512] with ReLU as the activation function. The function $g$ is modeled as a single fully connected layer.

\textbf{Task prediction module}.
As proposed in \cite{koh2020concept} and \cite{marconato2022glancenets}, we utilize a single fully connected layer mapping from the concepts to predictions ($h$). The final output is passed through a softmax layer to compute probability scores for the label. 


\subsection{Evaluation Setup}
\textbf{Task and concept accuracy.} 
For a Concept-Based Model to be deemed effective, it is required to be at par on performance with non-inherently explainable models (or black box models). We measure the task accuracy and compare it against a standard Deep Neural Network formed using the same encoder as the VAE and a fully connected layer as the classification head.

Next, we compare the concept-accuracy of CoLiDR against CBMs as proposed in \cite{koh2020concept}. For datasets with binary concepts, accuracy is reported as 0-1 error (misclassification). For datasets with non-binary concepts, Root Mean Square Error (RMSE) is reported. 
\\

\textbf{Disentangled representation aggregation performance.}
Effective aggregation of disentangled representations serves as the most important desiderata for CoLiDR. However, it is not straightforward to understand the aggregation effect. Unlike supervised disentanglement where each dimension of the disentangled representations is forced to correspond to a concept, CoLiDR aggregates multiple dimensions into a concept. Hence, instead of identifying individually significant dimensions as concepts, it is important to consider a set of representative dimensions from the aggregation module. 

Due to VAE's inherent disentanglement procedure, we can safely assume the mutual independence of dimensions among the learned disentangled representations. Hence, the latent dimensions themselves can be thought of as features of the aggregation module. In effect, the problem of selecting a representative set of dimensions is identical to assigning importance scores to the most important features in a deep neural network. Multiple post-hoc interpretability methods are proposed \cite{sundararajan2017axiomatic,ribeiro2016should}. We utilize \textit{Integrated Gradients (IG)}\cite{sundararajan2017axiomatic} to assign importance scores to each dimension. For an input $x$ and a baseline $x'$ (zero-vector), IG computes attribution scores for each feature $i$ using the following path integral:
\begin{equation}
    IG(x) = (x_i - x'_i) \cdot \int^1_{\alpha=0} \frac{\partial F (x' + \alpha \cdot (x - x'))}{\partial x_i} d\alpha
\end{equation}

We utilize Captum (\url{https://captum.ai/}) to compute attributions using IG. The values of the attributions are normalized and their absolute values are assigned as the final attributions. Once the attribution scores are computed, we utilize three distinct methods to evaluate the effects of the representative set of disentangled representations.
\begin{itemize}[leftmargin=*, parsep=0pt, itemsep=0pt, topsep=0pt]
    \item \textbf{GradCAM \cite{selvaraju2017grad} visualizations:} For each dimension in the top-k most important dimensions, the GradCAM attribution visualization plots are plotted. 
    \item \textbf{Latent space traversal:} For concepts with one dominant dimension in the representative set, we linearly interpolate the normalized latent space between two terminal values to identify visual cues in each generated image
    \item \textbf{Oracle Classification:} Even though the aforementioned visualizations provide qualitative evaluation, a robust quantitative evaluation is required to assuage the effects of confirmation bias. To achieve this, we train a \textit{Oracle Network} to automatically classify images to the associated concepts with high accuracy. Subsequently, we compare heatmaps generated using CBMs and CoLiDR with the actual ground-truth annotated bounding boxes and report the average Intersection over Union (IoU) scores which measure the ratio of overlap to the combined are between two bounding boxes (Refer Appendix for mathematical formulation). We utilize a fine-tuned VGG-16 model as the backbone of the oracle for each dataset. For more details, refer to Appendix.
    \\
\end{itemize}

\textbf{Concept decomposition performance (intervention).}
Another desirable desideratum of a Concept-based model is its ease of debugging. For a misclassified sample, fixing the concept annotation by a domain expert should be able to correct the model predictions. Recall that the decomposition module decomposes concepts back into disentangled representations. We define an intervention to be successful for a wrongly classified sample, replacing the predicted concept score with the ground truth concept annotations changes the wrong prediction label to the correct ground truth label $y$.

\subsection{Evaluation Metric Descriptions}
\subsubsection{Oracle Training}
We utilize a fine-tuned VGG-16 which contains 5 blocks of convolutional layers followed by a max pooling layer at the end of each block. The final output is passed through a 3-layer fully connected network to perform concept prediction. The model is trained similarly to the concept network where each concept is weighted by its relative occurrence in the dataset.

\subsubsection{Intersection over Union Calculation (IoU)}
Intersection over Union is defined as the ratio of overlap between two spans of areas on an image. Assuming two bounding boxes $A$ and $B$, IoU is defined as:
\begin{equation}
    \text{IoU} = \frac{A \cap B}{A \cup B}
\label{eq:iou}
\end{equation}
We calculate IoU's between heatmaps generated using GradCAM. The selected pixels above a threshold are selected (255 value grayscale CAM maps above 150). As the heatmaps can be irregular in shape, we consider each selected pixel as part of a set on a fixed image size.

\section{Results and Discussion}

\subsection{Task and Concept Accuracy}
In Table~\ref{tab:performance-compare}, we compare our model against Concept Bottleneck Models (CBMs) \cite{koh2020concept}, which are trained without disentanglement learning, and GlanceNets \cite{marconato2022glancenets}, which involves the learning of disentangled factors, on the average task performance and concept errors. We implement CBMs by replacing the VAE with a standard Autoencoder and the task learning module and the Aggregation/Decomposition module are replaced by identity functions. For GlanceNets, we provide supervision on not only the learned concepts but also a part of disentangled latent space (represented as $\mathbf{G_V}$ in \cite{marconato2022glancenets}). Differently, we do not utilize the Open-Set Recognition Mechanism as we do not specifically study concept leakage.

As can be seen in Table~\ref{tab:performance-compare}, CoLIDR variants perform the best among all concept-based models with disentanglement learning and even outperforms CBM on 3 out of 4 datasets. The results show that CBM performs better than GlanceNet and CoLiDR on the AWA2 dataset (0.531). A possible reason for this observation stems from the fact that disentanglement performance on the AWA2 dataset is not effectively captured by VAEs (Refer to Appendix) due to less training data and significant noise in the dataset. Furthermore, the performance of GlanceNet and CoLIDR is at par on dSprites and Shapes3D implying that datasets that are `easier' to disentangle yield similar performances across methods. 

For the concept identification task, CoLiDR still achieves the best performance with the lowest concept errors among all models with disentanglement learning. Compared with the model without disentanglement learning, Table~\ref{tab:performance-compare} shows that our model performs comparable and sometimes better than CBM on dSprtes, Shapes3D, and CelebA datasets. 
For the AWA2 datasets, CBM outperforms both CoLiDR and GlanceNet. We attribute this to the fact that CBM can learn concepts very flexibly without the regularization of disentanglement. However, this also leads to the problem of learning spurious correlation instead of real semantics of concepts, as shown in Appendix. Note that CoLiDR with no consistency regularization ($\mathcal{L}_{drc}$) performs the worst on concept learning - implying strong disentanglement performance is vital to concept learning. 


\subsection{Disentangled Representation Visualizations}
\begin{figure}[h]
    \centering
    \includegraphics[width=0.45\textwidth]{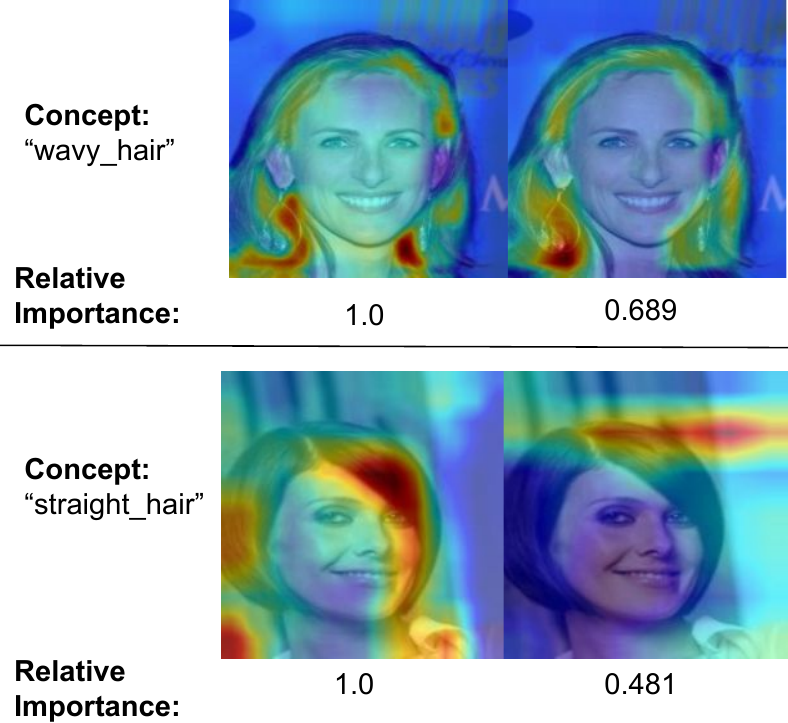}
    \caption{GradCAM visualizations of the top-2 highest attributed dimensions with respect to correctly predicted concepts ``wavy\_hair'' (top) and ``straight\_hair'' (bottom) on 2 distinct samples from CelebA.}
    \label{fig:viz-celeb}
\end{figure}

\begin{figure}[h]
    \centering
    \includegraphics[width=0.45\textwidth]{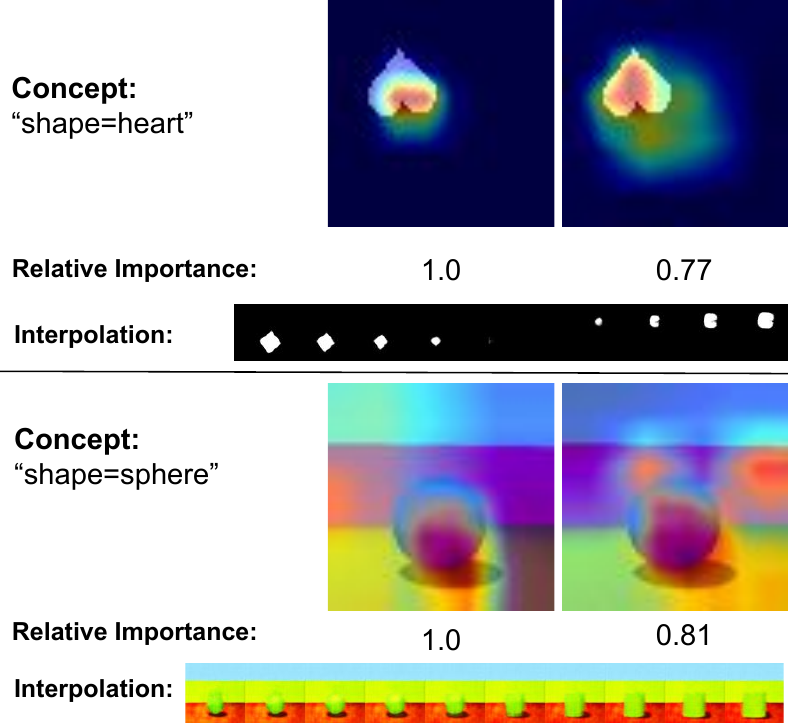}
    \caption{GradCAM visualizations of the top-2 highest attributed dimensions for the correctly predicted concept of shape for (top) d-sprites and (bottom) Shapes3D. The interpolation along the highest contributing dimension shows that the dimension effectively captures the shape of the object in the image.}
    \label{fig:viz-dsprites}
\end{figure}

Figures~\ref{fig:viz-celeb} and \ref{fig:viz-dsprites} respectively demonstrate the most important dimensions constituting a concept. In the Figure~\ref{fig:viz-celeb}, we demonstrate the concept ``brown\_hair'' and its associated dimensions with scores calculated using IG. For example, in Figure~\ref{fig:viz-celeb} GradCAM heatmap visualizations on the two highest computed constituent dimensions for a correctly identified concept ``straight\_hair'' (top) and ``wavy\_hair'' are shown. We see that both dimensions correctly correspond to the distributions constituting the hair of the people in the image. Similarly, in Figure~\ref{fig:viz-dsprites}, the GradCAM visualizations corresponding to the concepts ``heart'' (left) and ``ellipse'' (right) are shown. In addition, we also provide a linear interpolation of the identified dimensions. As can be seen, the dimensions are indeed responsible for the shapes of hearts and ellipses.

\subsection{Comparision with Oracle Network}
We partition a subset of concepts that can be easily understood in the CelebA dataset and dub them `simple' (Refer to Appendix for the exact splits). We report average IoU values for correctly classified concepts on both `simple' and all concepts in Table~\ref{tab:oracle}. Column-2, 3 and 4 list the average IoUs for simple concepts across CBMs, CoLiDR with only the top-2 and top-5 dimensions respectively.  Heatmaps produced by CBMs \cite{margeloiu2021concept} show very low IoU values implying they are not able to capture the spatiality of concepts well. On the other hand, CoLIDR with $\beta$-VAEs give much higher IoUs than CBMs - demonstrating that CoLiDR captures the spatial nature of concepts extremely well. The more dimensions visualized (two vs five), the lower the IoU goes - implying that most of the spatial concept information is encoded in only a few dimensions. 

\begin{table}[h]
\centering

\begin{tabular}{c|c|c|c|c}
\hline
                     & \textbf{Oracle} & \textbf{CBM/CEM} & \textbf{\begin{tabular}[c]{@{}c@{}}CoLiDR\\ (dim=2)\end{tabular}} & \textbf{\begin{tabular}[c]{@{}c@{}}CoLiDR\\ (dim=5)\end{tabular}} \\
\hline
`Simple' IoU  & 0.99   &  0.31  &  0.74   &  0.71  \\
Overall IoU & 0.99   &  0.24  &  0.63   &  0.62  \\                   
\hline                    
\end{tabular}

\caption{IoU values for GradCAM heatmaps generated using CBMs (Column-2) as compared to CoLiDR utilizing the top-2 dimensions (Column-3) and top-5 dimensions (Column-4). }
\label{tab:oracle}
\end{table}

\subsection{Test-time Intervention}
\begin{figure}[h]
    \centering
    \includegraphics[width=0.4\textwidth]{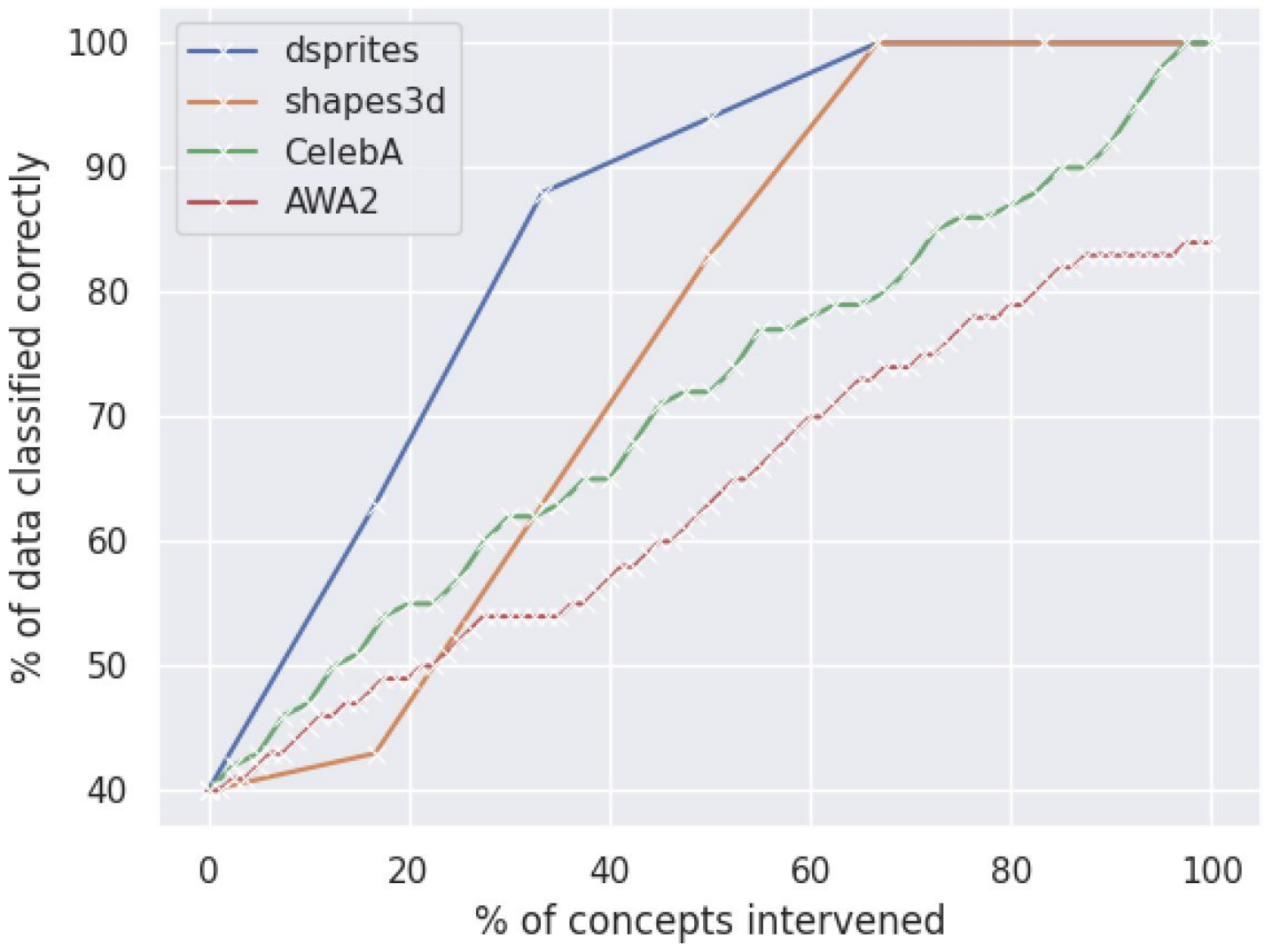}
    \caption{Test-time intervention for all datasets. CoLiDR Percentage of samples correctly classified increases after intervention as the percentage of concepts increases. }
    \label{fig:intervention}
\end{figure}

Figure~\ref{fig:intervention} shows the effect of intervening on a wrongly predicted concept by its ground truth value. As the number of concepts vary for each dataset, we plot the percentage of samples getting corrected after intervention (y-axis) v/s percentage of concepts replaced by their ground truth values (x-axis). We see that intervention is more successful on datasets with fewer concepts.

Additional results can be found in the Appendix.

\begin{figure}[h]
    \centering
    \fbox{\includegraphics[width=0.45\textwidth]{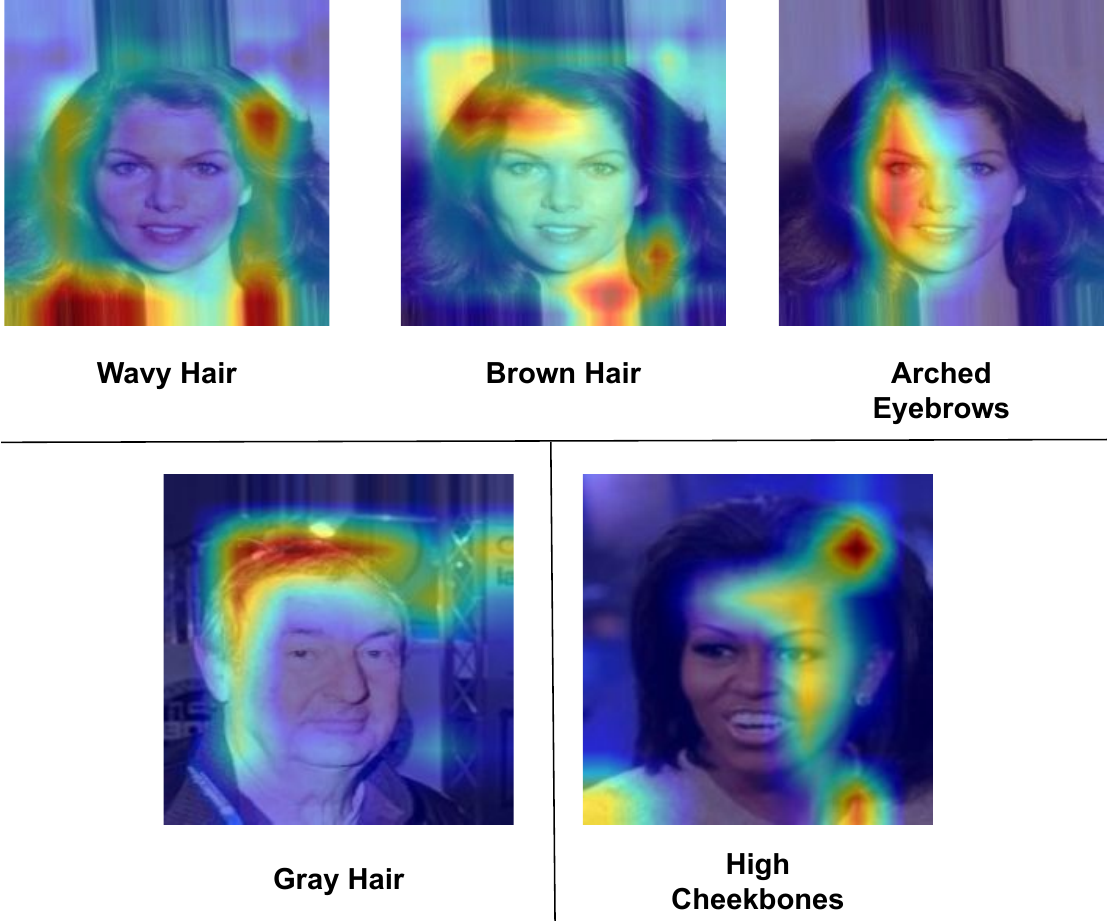}}
    \caption{GradCAMs Heatmaps corresponding to the most important dimensions of correctly predicted concepts of 3 distinct images belonging to test set of CelebA.}
    \label{fig:celeba-viz2}
\end{figure}
\section{Future Work}
CoLiDR is a highly generalizable framework that can be plugged into any disentanglement architecture. In this paper we demonstrate our results on Variational Autoencoders, however many more complex architectures can be utilized for learning disentangled representations. Generative Adversarial Networks are another class of stochastic non-linear latent representation learning framework which can be directly utilized as a DRL module. Another avenue of exploration can go along the lines where concepts instead of being independent of each other, are causally related - similar to \cite{heskes2020causal}. 

\section{Conclusion}
In this paper, we propose \textit{CoLiDR}, a novel interpretable concept-based model that learns concepts using aggregated disentangled representations. Empirical results demonstrate that CoLiDR bridges the gap between concept learning and disentangled representation learning by formatting human-understandable concepts as aggregations of fundamental generative factors. The performance of CoLiDR for task and concept prediction is on par with the concept-focused models (CBM/CEM) that lack the explainability of generative factors. CoLiDR can flexibly learn complex concepts as an aggregation of disentangled generative factors which improves its performance on both task and concept learning. Using an aggregation step, concepts learned using CoLiDR can be constituents of multiple generative factors, giving a much finer overview of interpretability.

\section*{Acknowledgements}
This work is supported in part by the US National Science Foundation under grants 2217071, 2213700, 2106913, 2008208, 1955151. Any opinions, findings, and conclusions or recommendations expressed in this material are those of the
author(s) and do not necessarily reflect the views of the National Science Foundation.

\bibliographystyle{ACM-Reference-Format}
\balance
\bibliography{main}

\appendix
\clearpage
\setcounter{page}{1}

\section{Appendix}
\label{sec:appendix}

\subsection{Downstream Task Descriptions}
\subsubsection{D-Sprites/Shapes3D}
We create 6 distinct binary tasks for the D-Sprites dataset. Each task is created by splitting the training set based on two factors. As the training set is randomly sampled, the number of training points in each task is different. However, the training set for each task is ensured to have balanced labels. 

\subsubsection{CelebA}
For CelebA dataset, we utilize the labels formed by clustering all the images into 10 clusters using the clustering algorithm. The choice of the number of clusters are made using TSNE visualization of the actual clusters. It is worth noting that we utilize ten clusters instead of four in \cite{marconato2022glancenets}. This important change in task is made due to the fact that we utilize a different subset of concepts (`simple' concepts) to model CoLiDR. Table~\ref{tab:celeb-concept} lists the set of concepts for which annotations are provided. We compute all the experimental results on the simple set of concepts (left). The simple concepts are chosen as they are objective in nature and can be easily discerned by visual inspection of heatmaps.  

\begin{table*}[h]
\begin{tabular}{c|c}
\hline
\textbf{Simple Concepts (20)} &  \textbf{All Concepts (40)} \\
 \hline
Arched\_Eyebrows,
Bangs, 
Big\_Lips,  & 5\_o\_Clock\_Shadow,
Arched\_Eyebrows,
Attractive,
Bags\_Under\_Eyes, \\ 
Black\_Hair, Blond\_Hair,
Brown\_Hair,
 & Bald,
Bangs,
Big\_Lips,
Big\_Nose,
Black\_Hair,
Blond\_Hair,
Blurry, \\
Double\_Chin,
Goatee,  Gray\_Hair,
 & Brown\_Hair,
Bushy\_Eyebrows,
Chubby,
Double\_Chin,
Eyeglasses\\
Male,
Mustache, Narrow\_Eyes,
Rosy\_Cheeks,
 & Goatee,
Gray\_Hair,
Heavy\_Makeup,
Male,\\
Sideburns,
Smiling, Straight\_Hair, & Mouth\_Slightly\_Open,
Mustache,
Narrow\_Eyes,
No\_Beard,
Oval\_Face, \\
Wavy\_Hair,
Wearing\_Earrings, Wearing\_Hat & Pale\_Skin,
Pointy\_Nose,
Receding\_Hairline,
Rosy\_Cheeks,
Sideburns,\\
High\_Cheekbones & Smiling,
Straight\_Hair,
Wavy\_Hair,
Wearing\_Earrings,
Wearing\_Hat,\\
 & Wearing\_Lipstick,
Wearing\_Necklace,
Wearing\_Necktie,
Young \\
\hline
\end{tabular} 
\caption{List of Simple Concepts (left) and the entire concept set (Right) for the CelebA dataset}
\label{tab:celeb-concept}
\end{table*}

\subsubsection{AWA2}
For AWA2 dataset, we keep the task the same as the identity classification of 50 different identities. The original dataset is annotated with 85 different binary concepts. The list of concepts can be found \footnote{\url{https://cvml.ist.ac.at/AwA2/}}. For our experiments, we utilize a smaller subset of 30 concepts. The subset is specifically chosen to only capture concepts which are objective and human-understandable.


\subsection{Implementation Details}
\subsubsection{Model Architecture}
\noindent \textbf{Disentangled Representations Learning (DRL) Module:}
Table~\ref{tab:model-arch} details the major components of both architectures respectively. The standard VAE suffers from the well-documented problem of \cite{higgins2018beta} over-regularization of the disentangled representations $\mathbf{z}$, which hinders the quality of visualized reconstructions $\mathbf{\hat{x}}$ due to the information constraint imposed by the independent Gaussian priors. Hence, we evaluate CoLiDR on both a standard VAE with $\beta = 1$ and a $\beta$-VAE, where the values of $\beta$ are tuned separately on each dataset. Specifically, we employ the following values: dSprites: $\beta=0.025$, Shapes3D: $\beta=0.025$, CelebA: $\beta=2.5e-5$ and AWA2: $\beta=1e-5$.

\noindent \textbf{Aggregation/Decomposition Module:} The Aggregation module $\mathcal{A}$ maps between the latent representations $\mathbf{z}$ and concept set $\mathbf{C}$  while the Decomposition module $\mathcal{D}$ maps the concept set $\mathbf{C}$ to the latent representations $\mathbf{z}$ using a set of non linear transforms $\{a_i\}$ and $\{d_i \}$ where i=\{0,..k\}. We model each transformation as a fully connected network mapping from a single dimension in the latent space ($\mathbf{z}$) to the transformed latent space ($\mathbf{z'}$) and vice-versa. The number of the intermediate layers is 2 for dSprites/Shapes3D each of size 512, and the number of the intermediate layers is 4 for CelebA/AWA2 each of size 512. The map between the transformed latent space $z'$ and $C$ and vice versa - is modeled as a single feed-forward layer.

\noindent \textbf{Task Prediction Module:}
The task prediction module utilizes a linear fully connected layer mapping from the annotated concepts (Row-5 in Table~\ref{tab:model-arch}) to the number of task labels (Row-7 in Table~\ref{tab:model-arch}). The final predictions are performed after converting the model outputs into probabilities using the softmax operator. 

\subsubsection{Training Hyperparameter Settings}

\noindent{\textbf{Sequential Training:}}
To avoid posterior collapse, we first train the VAE without the influence of any other loss term, i.e., $\lambda_1,\lambda_2,\lambda_3,\lambda_4 = 0 $, only utilizing the ELBO term (Equation~\ref{eq:final-obj}). Subsequently, we only learn the concept mappings, i.e., $\lambda_2 = 0$ as proposed in Sequential training of  \cite{koh2020concept}. 
Finally, we train the entire model architecture end-to-end using carefully tuned hyperparameters $\lambda_1,\lambda_2,\lambda_3,\lambda_4$ for each dataset. The detailed values are given below:
\begin{itemize}
    \item dSprites: $\lambda_1=0.5,\lambda_2=0.2,\lambda_3=1,\lambda_4=0.1$
    \item Shapes3D: $\lambda_1=0.5,\lambda_2=0.2,\lambda_3=1,\lambda_4=0.1$
    \item CelebA: $\lambda_1=10,\lambda_2=0.08,\lambda_3=0.1,\lambda_4=0.01$
    \item AWA2: $\lambda_1=5,\lambda_2=0.05,\lambda_3=0.1,\lambda_4=0.01$
\end{itemize}
\noindent \textbf{Learning Rate and Scheduling:} Training VAEs are notoriously highly sensitive to the LR and optimizer. We utilize the Adam Optimizer with an initial learning rate of 5e-3 for dSprites/Shapes3D, 1e-3 for CelebA and 5e-4 for AWA2. The batch sizes are 64 each for dSprites and Shapes3D, while it is set at 32 for CelebA and 16 for AWA2. We train the entire model architecture for 100 epochs with the first 50 epochs training the VAE, the subsequent 25 epochs training the Aggregation/Decomposition module while the last 25 epochs train the entire model, end-to-end. 



\subsection{Additional Visual Results}
\begin{figure}[t]
    \centering
    \includegraphics[width=0.45\textwidth]{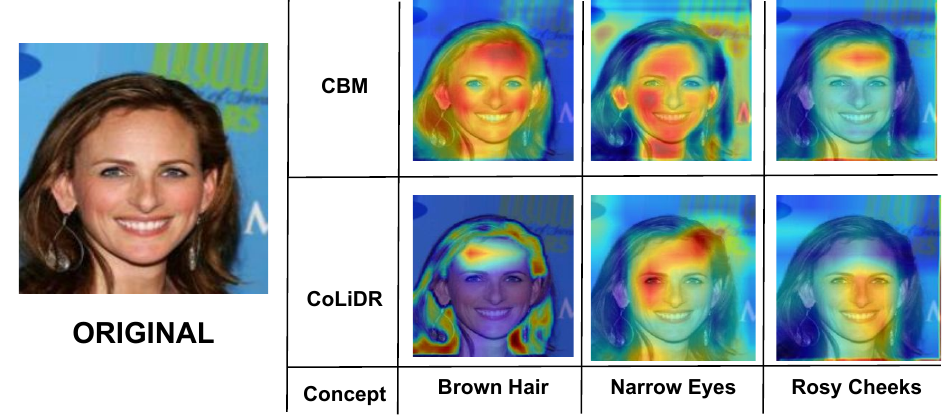}
    \caption{GradCAMs Heatmaps corresponding to a test image from CelebA. The Top Row shows heatmaps computed on CBMs while the bottom row shows the heatmaps corresponding to the most important dimensions of correctly predicted concepts using CoLiDR. As can be seen: heatmaps from CoLiDR capture the spatiality of concepts much better than CBMs.}
    \label{fig:celeba-viz1-app}
\end{figure}

\begin{figure*}[h]
    \centering
    \begin{subfigure}{.48\textwidth}
  \centering
  \includegraphics[width=0.6\textwidth]{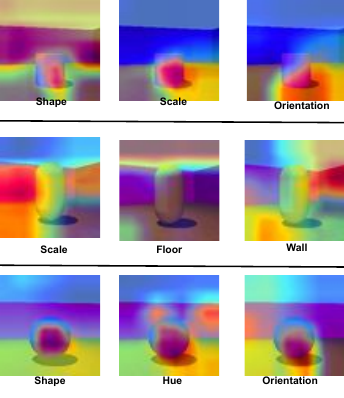}
  \caption{GradCAMs Heatmaps corresponding to the most important dimensions of correctly predicted concepts of 3 distinct images belonging to test set of Shapes3D.}
  \label{fig:shapes-viz1}
\end{subfigure}
\begin{subfigure}{.48\textwidth}
  \centering
  \includegraphics[width=0.9\textwidth]{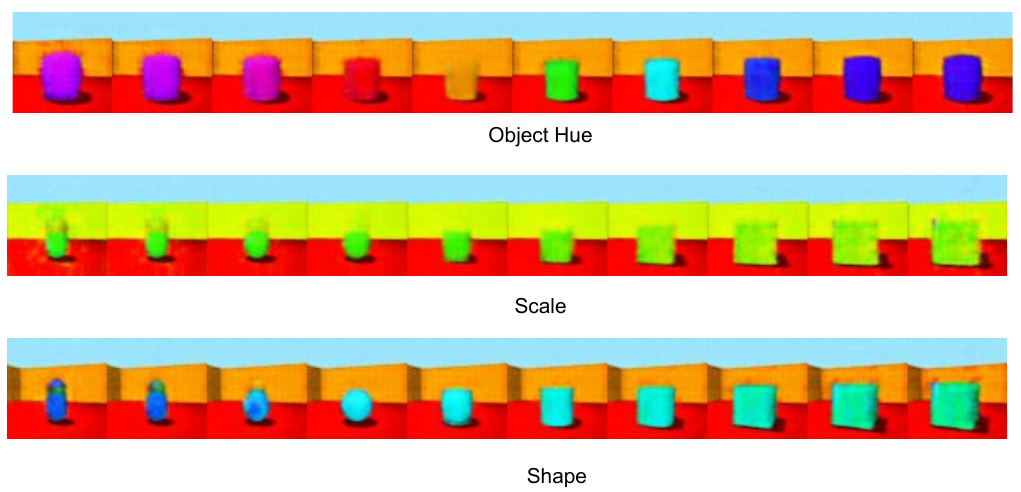}
    \caption{Linear interpolation of randomly sampled latent space corresponding to images. The interpolated space corresponds well with the concepts it contributes to the most.}
  \label{fig:shapes-viz2}
\end{subfigure}
\caption{Additional results on the Shapes3D dataset}
    
\end{figure*}

Figure~\ref{fig:celeba-viz1-app} demonstrates the highest attributed latent dimensions associated with correctly classified concepts. In particular. we compare the heatmaps generated using CoLiDR with the heatmaps generated using CBMs. As can be seen in Figure~\ref{fig:celeba-viz1-app}, the heatmaps corresponding to the most important dimensions of blond hair, narrow eyes, and rosy cheeks correspond to their respective correct positions in the image using CoLiDR as compared to CBMs. Similarly, Figure~\ref{fig:shapes-viz1} shows the heatmaps corresponding to correctly classified concepts from the Shapes3D dataset. In addition, in Figure~\ref{fig:shapes-viz2}, we provide linear interpolation of the most contributing latent dimensions corresponding to a concept.

\begin{table*}[t]
\centering
\begin{tabular}{c|c|c}
\hline
Component    & \textbf{dSprites/Shapes3D} & \textbf{CelebA/AWA2} \\
\hline
\textbf{Encoder}   &  3x & 5x \\
& {[} Conv2D(filters, kernel=3, stride=2); &  {[} Conv2D(filters, kernel=3, stride=2)\\
& BatchNorm2D(); & BatchNorm2D();\\
& LeakyReLU(){]} & LeakyReLU(){]}\\ 
& filters in {[}32, 64, 64{]} & filters in {[}32, 64, 128, 256, 512{]} \\ 
\hline
\textbf{Latent Dim} & 64  & 512         \\
\hline
\textbf{Decoder}   &  3x & 5x \\  
& {[} ConvTranspose2D, filters, kernel=3, stride=2; & {[} ConvTranspose2D, filters, kernel=3, stride=2;\\
& BatchNorm2D(); & BatchNorm2D(); \\ 
& LeakyReLU(){]} & LeakyReLU(){]}\\ 

& filters in {[}64, 64, 32{]} & filters in {[}512, 256, 128, 64, 32{]} \\ 
\hline

\textbf{Aggregation}   &  64x [Linear(1,64); ReLU();Linear(64,64); ] & 512x [Linear(1,512); ReLU(); Linear(512,512);\\
& ReLU(); Linear(64,1) & ReLU(); Linear(512,1)] \\
\hline

\textbf{Concepts} & Annot = 6; Total = 64 & Annot = 20 (CelebA), 30 (AWA2); Total = 512\\

\hline 

\textbf{Decomposition}   &  64x [Linear(1,512); ReLU(); Linear(512,1)] & 512x [Linear(1,512); ReLU(); Linear(512,512);\\
& & ReLU(); Linear(512,512)] \\

\hline

\textbf{Task Labels} & 2 & 10 (CelebA), 50 (AWA2) \\
\hline

\end{tabular}

\caption{CoLiDR module architecture details for the dSprites, Shapes3D datasets (Left) and  CelebA, AWA2 datasets (Right). We utilize identical Encoder and Decoder stacks and identical input/output image sizes across all datasets.}
\label{tab:model-arch}

\end{table*}




\end{document}